\documentclass{article}

\usepackage[nonatbib, preprint]{neurips_2022}

\usepackage[utf8]{inputenc} 
\usepackage[T1]{fontenc}    
\usepackage{hyperref}       
\usepackage{url}            
\usepackage{booktabs}       
\usepackage{amsfonts}       
\usepackage{nicefrac}       
\usepackage{microtype}      
\usepackage{xcolor}         
\usepackage[acronym]{glossaries}
\newacronym{ai}{AI}{Artificial Intelligence}
\newacronym{dnn}{DNN}{Deep Neural Network}
\newacronym{cnn}{CNN}{Convolutional Neural Network}
\newacronym{gan}{GAN}{Generative Adversarial Network}
\newacronym{vae}{VAE}{Variational AutoEncoder}
\newacronym{erm}{ERM}{Empirical Risk Minimization}
\newacronym{rnn}{RNN}{Recurrent Neural Network}
\newacronym{mlp}{MLP}{Multi-Layer Perceptron}
\newacronym{map}{MAP}{Mean Average Precision}
\newacronym{sis}{SIS}{Segmentation Inconsistency Score}
\newacronym{sil}{SIL}{Segmentation Inconsistency Loss}
\newacronym{iid}{IID}{Independent and Identically Distributed}
\newacronym{ind}{InD}{In-Distribution}
\newacronym{iou}{mIoU}{Mean Intersection over Union}
\newacronym{fpn}{FPN}{Feature Pyramid Network}
\newacronym{ood}{OOD}{Out of Distribution}
\newacronym{sisauc}{SIS-AUC}{Segmentation Inconsistency Score Area Under Curve}
\newacronym{diou}{\(\Delta \% IoU\)}{Generalizability Gap}
\newacronym{cs}{C.StD}{Coefficient of Standard Deviation}
\usepackage{graphicx}
\usepackage{tabularx, cleveref}

\usepackage[backend=bibtex,style=numeric-comp]{biblatex}
\title{Segmentation Consistency Training: Out-of-Distribution Generalization for Medical Image Segmentation}

\author{%
  Birk Torpmann-Hagen \\
  University of Oslo\\
  \texttt{birk.torpmann.hagen@gmail.com} \\
   \And
   Vajira Thambawita \\
   SimulaMet \\
   \texttt{vajira@simula.no} \\
      \And
   Kyrre Glette \\
   University of Oslo \\
   \texttt{kyrrehg@ifi.uio.no} \\
    \And
    P{\aa}l Halvorsen \\
   SimulaMet \\
   \texttt{paalh@simula.no} \\
      \And
   Michael A. Riegler \\
   SimulaMet \\
   \texttt{michael@simula.no} \\
}

\begin{document}

\maketitle

\begin{abstract}
  Generalizability is seen as one of the major challenges in deep learning, in particular in the domain of medical imaging, where a change of hospital or in imaging routines can lead to a complete failure of a model. To tackle this, we introduce \textbf{Consistency Training}, a training procedure and alternative to data augmentation based on maximizing models' prediction consistency across augmented and unaugmented data in order to facilitate better out-of-distribution generalization. To this end, we develop a novel region-based segmentation loss function called \textbf{Segmentation Inconsistency Loss (SIL)}, which considers the differences between pairs of augmented and unaugmented predictions and labels. We demonstrate that Consistency Training outperforms conventional data augmentation on several out-of-distribution datasets on polyp segmentation, a popular medical task. 

\end{abstract}

\section{Introduction}
    The last decade or so has seen a veritable revolution in artificial intelligence. This has in large part been spearheaded by advancements in deep learning, the remarkable performance of which it can be argued has rendered more conventional approaches practically obsolete. Recent work has, however, highlighted that \glspl{dnn} are highly prone to exhibiting significant reductions in performance when deployed in practical settings or otherwise \gls{ood} data, in spite of the fact that they readily exhibit high performance when evaluated on previously unseen subsets of the training data~\cite{damour2020underspecification, shortcut_learning, noise_robustness, corruption_robustness}. This is referred to as \textit{generalization failure}. 
  
    Recent analyses attribute generalization failure to a structural misalignment between the features that a given model learns through \gls{erm} and the causal structure which it ideally should encode ~\cite{adversarial_bugs_features,shortcut_learning,IRM, causality}. Generally, this misalignment occurs as a result of the predictor - i.e., the trained model - learning spurious or otherwise causally unrepresentative features that nonetheless perform well within the training distribution. This is often referred to as \textit{shortcut learning}~\cite{shortcut_learning} or the \textit{Clever Hans effect}~\cite{cleverhans}. This behaviour is of course made evident as soon as the predictor is exposed to any form of distributional shift which breaks these shortcuts, at which point it will fail to generalize. These distributional shifts can range in magnitude, from common corruptions such as noise or blurs~\cite{corruption_robustness} or spatial transforms~\cite{spatial_robustness}, to practically imperceptible perturbations, typically exemplified by adversarial attacks~\cite{adversarial_attacks}, or as will be shown in this work; simply collecting data from different centers~\cite{gi_dataset_bias}. \gls{erm} does not and cannot guarantee invariance to these sorts of distributional shifts, as it assumes that the distribution of the training data is \gls{iid} to the true distribution~\cite{deep_learning_book}.
    
    Closely related to shortcut learning is underspecification~\cite{damour2020underspecification}. A machine learning pipeline can be considered underspecified when it can return any number of risk-equivalent predictors when evaluated on an \gls{iid} holdout set, dependent only on the random variables used within the training procedure - i.e., dropout, weight initialization, and so on. Even with identical hyperparameters, a given training procedure can return any number of predictors, each having learned different patterns within the dataset. One predictor may have learned one shortcut, another may have learned a different shortcut, and the next may actually have learned features that correspond to the causal structure it is intended to learn. With \gls{erm}, and in particular with \gls{ind}-oriented evaluation procedures, these are all erroneously considered equivalent.
 
    EndoCV2021 provided an opportunity to investigate generalization failure and means by which to counteract them in the context of detection- and segmentation of colorectal polyps via a competition~\cite{endocv2021}. Though several teams made good progress towards increasing generalizability, the organizers' review of the submissions~\cite{endocv2021_review} highlighted that every submitted model nevertheless exhibited significant performance reductions on the provided \gls{ood} datasets. Moreover, though a multitude of methods and approaches were tested, many of which did indeed benefit generalizability, few methods stood out as having the potential for significant further development. 
 
    To address these shortcomings, we introduce \textbf{Consistency Training}. We re-frame the problem of learning generalizable features into a matter of learning to \textit{not} learn spurious features. This framework requires a \textit{perturbation model}, which we in this work implement as simple data augmentation, and a differentiable quantity that represents the consistency of the predictions across perturbed and unperturbed inputs images, which we implement as \textit{\gls{sil}}, a Jaccard-like loss function that quantifies the degree to which the segmentation probability maps exhibit unwarranted change after the input is perturbed. This loss function is then used in conjunction with a task-specific loss, in this work Jaccard loss. To increase the stability of the training routine, we also implement a dynamic weighting procedure for the two constituent components of the overall loss function.  
    We show that Consistency Training increases generalization by a significant margin on all tested datasets when compared to conventional data augmentation. This framework is in other words a more performant alternative to data augmentation. Consistency Training leads to increased generalization with no additional overhead asides from the added computational cost involved with computing the auxiliary loss term and the memory required to store augmented and un-augmented versions of each batch.
    We summarize our contributions as following:

\begin{itemize}
    \item We introduce \gls{sil}, a novel region-based segmentation loss function  which quantifies the inconsistency between two predicted segmentations when the inputs are subjected to arbitrary augmentations.
    \item We propose a robust method of incorporating this loss function without a loss of segmentation performance through a dynamic weighting method.
    \item We demonstrate quantitatively that Consistency Training increases generalization when compared to data augmentation on three \gls{ood} datasets. 
\end{itemize}

\section{Related Work}
\textbf{Generalization Failure. }
The development of consistency training was in large part informed by recent advances in the understanding of generalization failure. D'Amour et al.~\cite{damour2020underspecification} perform a thorough analysis of generalization failure through multiple case studies and highlight the role of underspecification therein. Geirhos et al.~\cite{shortcut_learning} explore the idea of shortcut learning in a similar manner, and highlight the importance of learning causally related features. Schölkopf~\cite{causality} discusses the importance of causality in machine learning and how it relates to generalization failure. 

\textbf{Generalizable Training Methods. }
Increasing generalizability is an open problem, and there exists a large diversity of different approaches and perspectives on the matter in the literature. Arjovsky et al.~\cite{IRM} develop a novel training paradigm that makes use of multiple training environments in order to increase generalization. Robey et al.~\cite{modelbased} employ a similar method and develop a model-based training paradigm which attempts to induce invariance to learned mappings between training environments. Sandfort et al.~\cite{cyclegan} also leverage generative networks, but instead simply use generated CT-images as data augmentation, which they show improves \gls{ood} performance. Gokhale et al.~\cite{generalization_datamod} compare the use of multiple data modification methods on robustness and generalization and find that data augmentation improves generalizability by a significant margin.  Finally, Hendrycks et al.~\cite{augmix} incorporate a consistency term into their loss function, in particular Jensen-Shannon distance between output probabilities - in order to facilitate robustness to distributional shifts for the image-classification task. 

\textbf{Generalizable Polyp Segmentation. }
In the context of polyp-segmentation, this work was motivated in large part by the findings in the proceedings of EndoCV2021~\cite{endocv2021}, which through the evaluation of submissions on multiple \gls{ood} datasets highlighted the significance of generalization failure. The winning submission to EndoCV2021, submitted by Thambawita et al.~\cite{divergentnets}, leverages an ensemble-network in order to increase generalizability. Honga et al.~\cite{endoensemble} also implement an ensemble-based model, which they show improves generalization. Gu et al.~\cite{attention_generalizability} make use of domain composition and attention in an attempt to generalize to unseen domains.

In this regard, the existing works do not work well in the context of generalizability. Therefore, we introduced a novel training method with a new loss function to improve the generalizability of segmentation models on out-of-distribution datasets.  

\section{Approach}
\subsection{Consistency Training Method}
\begin{figure}[htb]
    \centering
    \includegraphics[width=\linewidth]{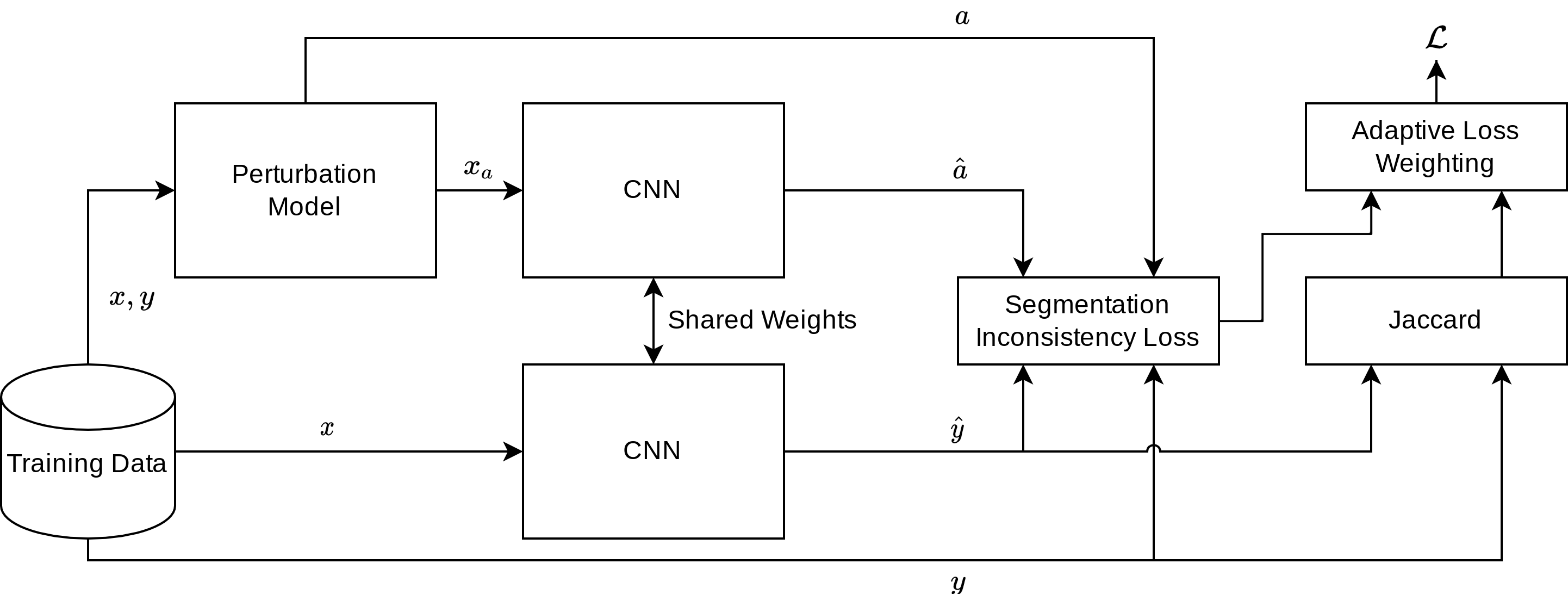}
    \caption{Diagram showing Consistency Training. The CNN is given two images, where one is simply an augmented version of the other. It then outputs two segmentations, which in conjunction with the labels for both images is used to compute \gls{sil}. The \gls{ind} \gls{iou} is then calculated, and used as weights for this term and a segmentation loss, in our case Jaccard loss. }
    \label{fig:consistency_training}
\end{figure}

This section will introduce Consistency Training, a training procedure wherein the objective is to optimize for invariance to a set of various image transformations by quantifying the degree to which the model outputs inconsistent predictions when its input is subjected to some transformations. This is achieved by giving the model two images: one which is augmented, and one which is not. These inputs are then passed through the model, resulting in two segmentation masks. The difference between these two predictions is then computed, and compared to the difference (if any) between the augmented and unaugmented segmentation labels. This is then incorporated into the loss-function such that the discrepancy between the expected prediction change and actual prediction change is minimized. This is illustrated in Figure \ref{fig:consistency_training}. The next sections will cover the theoretical basis of this training procedure as well as the implementation of its constituent components.

\subsection{Quantifying Segmentation Consistency}
Let \(Y:=\{y,\hat{y}:=f(x)\}\) be the set consisting of the segmentation labels (masks) and predictions for the unperturbed samples, where \(f(\cdot)\) as before denotes the segmentation model. Let \(\epsilon(\cdot)\) be some perturbation function. Then, let \(A:=\{a:=\epsilon(y),\hat{a}:=f(\epsilon(x))\}\) be the set consisting of masks and segmentation predictions  when the input is subjected to a perturbation. \textit{Segmentation Inconsistency} can then be quantified as:

\begin{equation}\label{eq:inconsistency}
    \overline{\mathcal{C}}(y, a, \hat{y}, \hat{a}) = \frac{\sum \{y\ominus\hat{y}\ominus a\ominus\hat{a}\}}{\sum\{y \cup a \cup \hat{y} \cup \hat{a} \}} 
\end{equation}
\(\ominus\) here denotes the symmetric difference/disjunctive union. Equivalently, \textit{Segmentation Consistency} can be expressed by:

\begin{equation}\label{eq:consistency}
    \mathcal{C}(y, a, \hat{y}, \hat{a}) = \frac{\sum \{y\cup\hat{y}\cup a\cup\hat{a}\} \ominus \{y\ominus\hat{y}\ominus a\ominus\hat{a}\}}{\sum\{y \cup a \cup \hat{y} \cup \hat{a} \}} 
\end{equation}
These formulations are related by:
\begin{equation*}
    \mathcal{C}(y, a, \hat{y}, \hat{a}) = 1-\overline{\mathcal{C}}(y, a, \hat{y}, \hat{a})
\end{equation*}

\begin{figure}[htb]
    \centering
    \includegraphics[width=\linewidth, trim={0 1.5cm 0 0},clip]{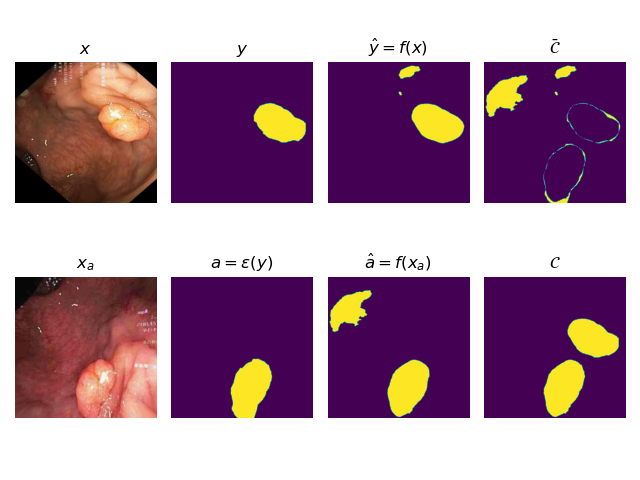}
    \caption[Segmentation Consistency Visualization 1]{Visualisation of the operations used when computing segmentation-consistency and -inconsistency. Segmentation Inconsistency \((\overline{\mathcal{C}}\) considers the ratio of pixel-predictions that underwent change as a result of a perturbation (here: augmentation), but should not have. Conversely, Segmentation Consistency \(\mathcal{C}\) considers the ratio of predictions that changed accordingly with respect to the perturbation.}
    \label{fig:consistency_example}
\end{figure}

In simple terms, this quantity corresponds to counting the number of pixels that change after the input is subjected to a perturbation,\(\hat{a}\ominus \hat{y}\), but discounting those we expect to change, \(a\ominus y\). This is shown in ~\Cref{fig:consistency_example}. 

Inconsistency as expressed in~\Cref{eq:inconsistency} is not differentiable, and thus it cannot in its current state be used as a part of a loss function. Thus, a smooth extension of this metric is needed which can be achieved in much the same way as how the Jaccard loss can be derived from the Jaccard index - i.e., by using differentiable versions of the set functions. 

We can extend the definition of the symmetric difference to \(\Theta(A,B) = A(1-B) + B(1-A)\). This, naturally, is equivalent to the standard symmetric difference if the values of A and B are binary. Similarly, the union operator can be extended as \( \bigcup(A,B) = A+B-AB\), and the intersection operator as \( \bigcap(A,B) = AB\). Like its binary equivalents, these operators maintain their associative and commutative properties. One can optimize for consistency by replacing the operators in~\Cref{eq:inconsistency} with these functions, which in turn can be used as a loss function:
\begin{equation}\label{inconsistency}
    L_{sil}(y, \hat{y},  a, \hat{a}) = \sum \frac{\Theta(y, \hat{y},  a, \hat{a})}{\bigcup(y, \hat{y},  a, \hat{a})}
\end{equation}
This loss function will from this point be referred to as \gls{sil}. Note that though this loss is implemented for binary segmentation in this task, it can be extended to multi-class segmentation by computing it for each channel and then reducing over channels. 

\subsection{Incorporating Consistency into Training}
  Using \gls{sil} as a loss function on its own is not really useful since it only expresses inconsistency, and is to a large extent agnostic to whatever object it is trying to segment. To illustrate, consider a model that predicts that every pixel is positive regardless of the content of the image, and that the augmentation strategy does not make use of augmentations that affect the labels. In this case, the consistency term will  always be zero. For example, if the augmentation being performed is simply additive noise, the inconsistency term is equally well minimized if the model learns to predict that every pixel is positive as it would be if the model learned to be robust to additive noise. Consequently, it has to be combined with a segmentation loss, for instance Jaccard loss. A simple way to do this would be to simply add them together and normalize, i.e.:
\begin{equation*}
    L(Y, A) = \frac{1}{2} \big[L_{seg}(Y)+L_{sil}(Y,A)\big]
\end{equation*}
Preliminary experiments showed that this, however, exhibited some degree of instability during training. The model would readily get stuck in local minima where its predictions were indeed consistent, but also consistently predicting artifacts. Examples of this can be found in the Appendix. 

To mitigate this, it is possible to employ a weighting strategy. Instead of simply adding the respective losses together, one may weight the individual components adaptively according to the \gls{ind} segmentation performance, for instance \gls{iou}. This way, the model will learn to predict generally correct segmentations early in the training, then start weighting consistency and as a result generalization more and more as the model sees improvements to its segmentation performance:
    \begin{equation}
        L = (1-IoU)\times L_{seg} + IoU \times L_c
    \end{equation}
Using this formulation, the model will start off trying to learn features that contribute to generally improved segmentation performance, then as segmentation performance improves start principally focusing on learning to be consistent. If the model starts veering into areas in the loss-landscape that constitute poor segmentation performance, it will self-correct by weighing the segmentation loss more. In the implementation used in this study, the \gls{iou} weights were calculated on a per-batch basis such that the model can quickly adapt if either of the respective objectives exhibit a degradation in performance during training.

\section{Experiments and Results}
To determine the generalizability of our methods, we trained ten instances each of four separate models using Consistency Training, as as well as with conventional data augmentation and no augmentation, which served as baselines. The generalizability of these models was then determined through computing \gls{iou} on three \gls{ood} datasets. The \glspl{iou} for models trained with Consistency training was then compared to the \glspl{iou} of the two baselines across the three \gls{ood} datasets to ascertain the extent to which it impacts generalization.

\subsection{Experimental Setup} \label{setup}
\textbf{Models. }
To evaluate the impact of Consistency Training sufficiently, it was tested across a range of different models.  These models include DeepLabV3+~\cite{deeplab}, \gls{fpn}~\cite{fpn}, UNet~\cite{unet}, and Tri-Unet~\cite{divergentnets}.

The models were implemented in pytorch using the segmentation-models-pytorch library~\cite{smp}, using the library's default values. This includes initialization with Imagenet-pretrained weights. Ten instances of each model were trained across each configuration in order to perform statistical analysis.

\begin{figure}[htb]
        \centering
        \begin{tabular}{cccc}
             Kvasir-SEG & Etis-LaribDB & CVC-ClinicDB &EndoCV2020 \\
             \includegraphics[width = 1in, height = 1in]{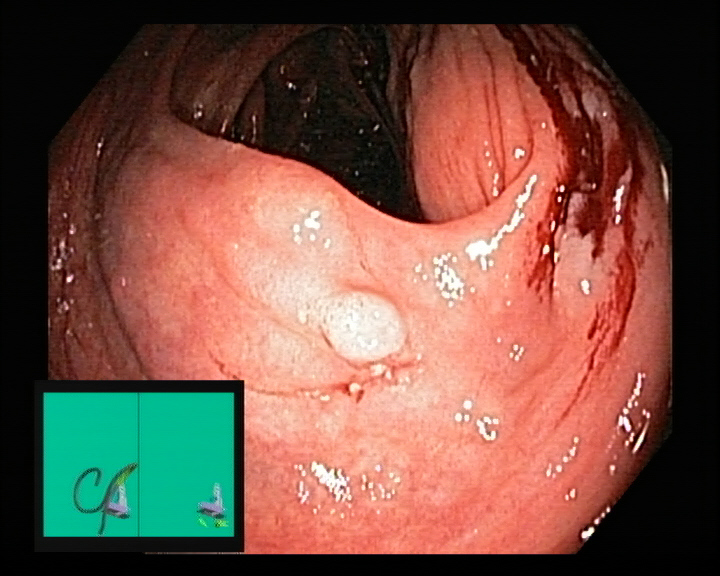} &   \includegraphics[width = 1in, height = 1in]{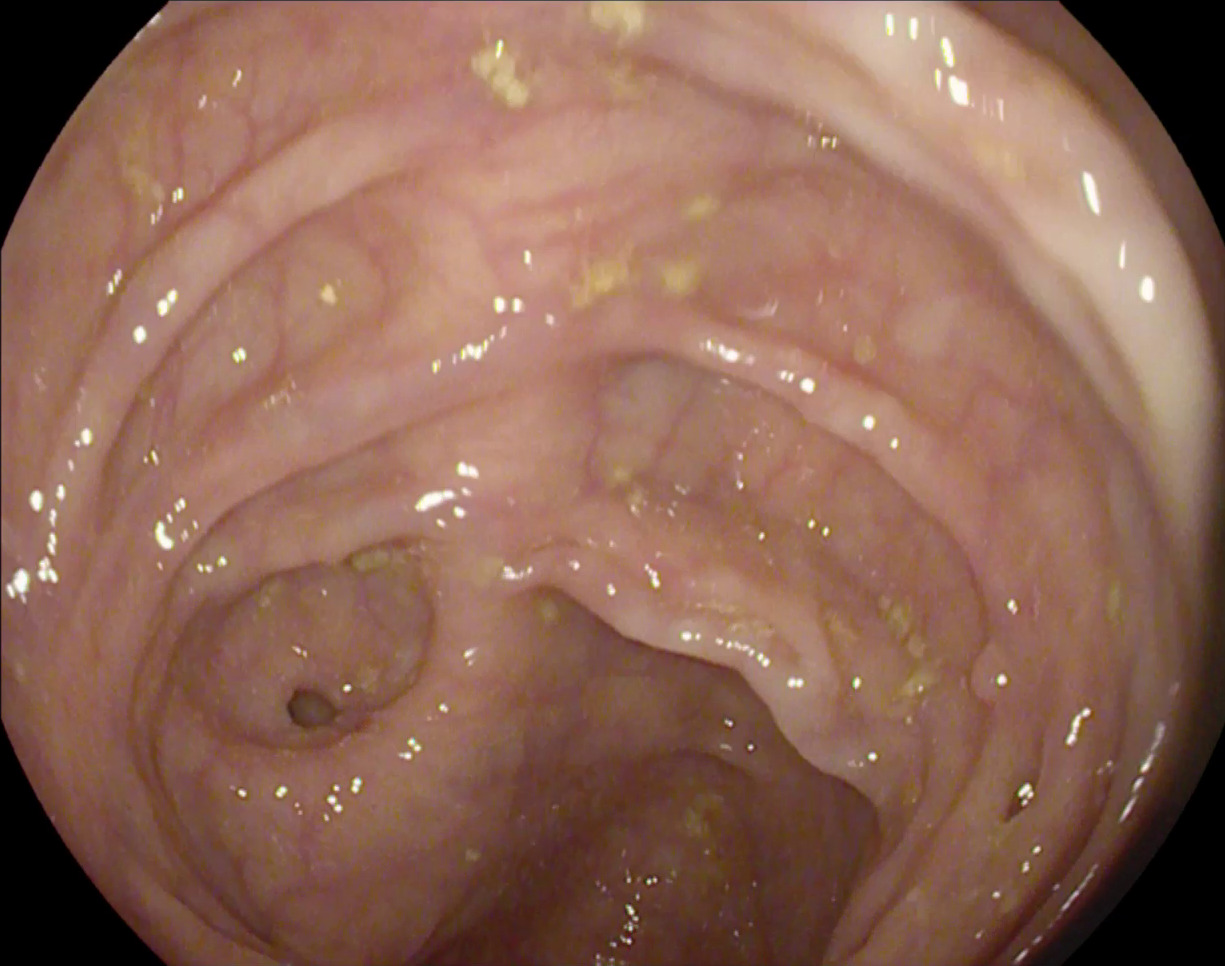} & \includegraphics[width = 1in, height = 1in]{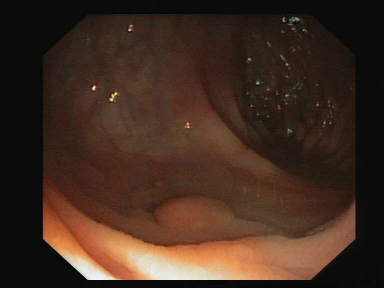} & \includegraphics[width = 1in, height = 1in]{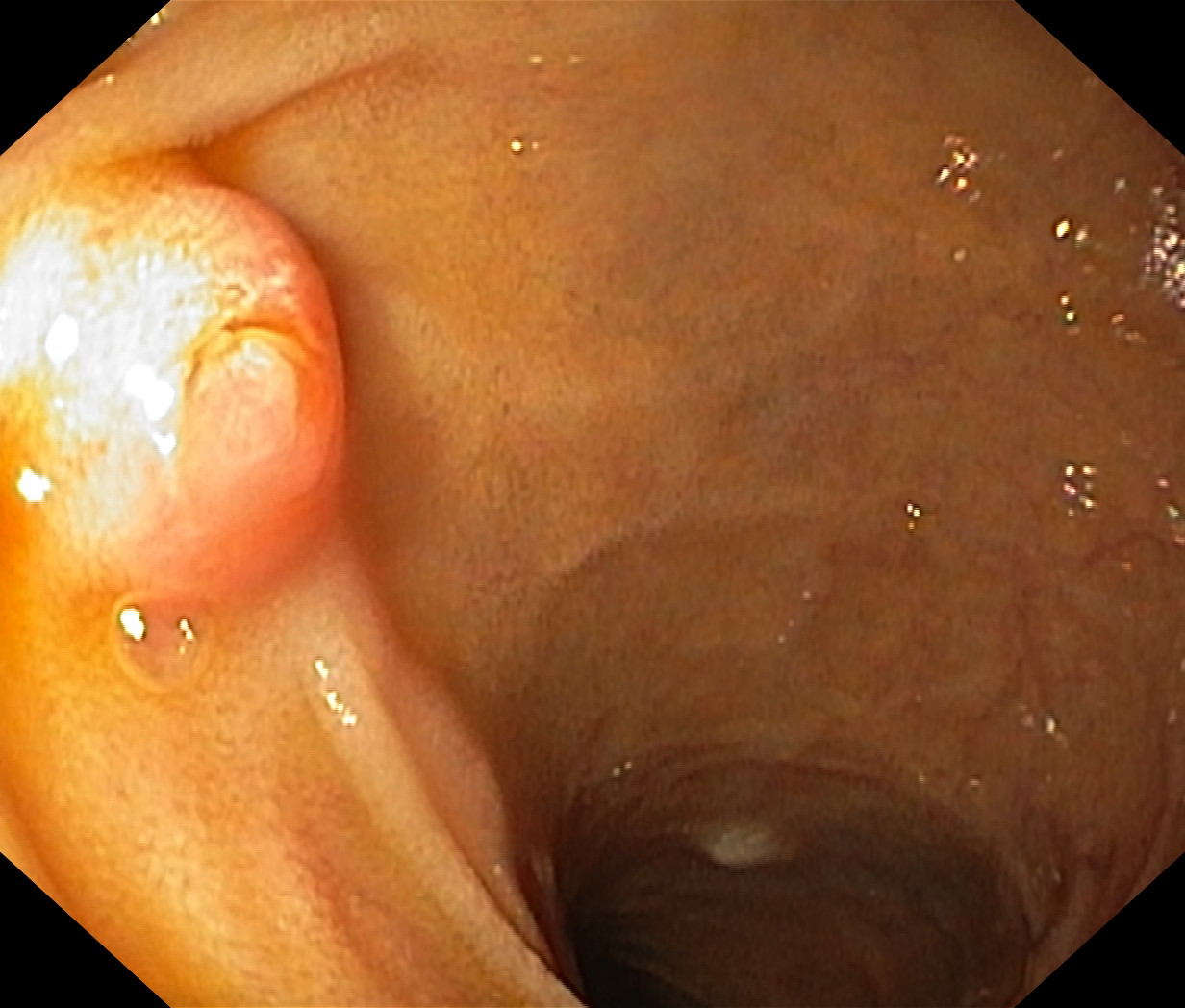} \\
              \includegraphics[width = 1in, height = 1in]{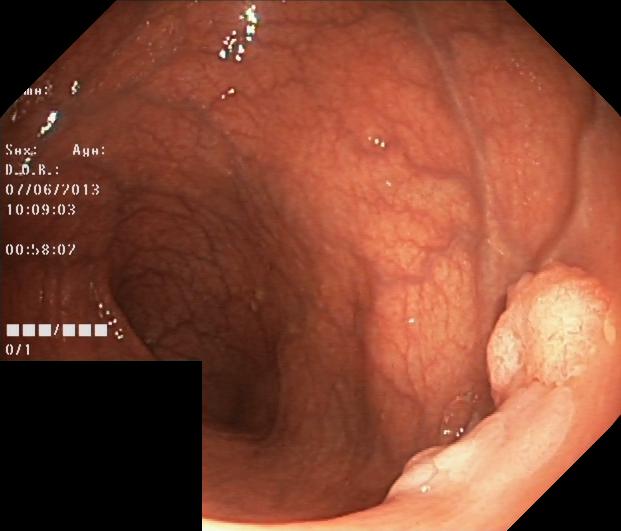} &   \includegraphics[width = 1in, height = 1in]{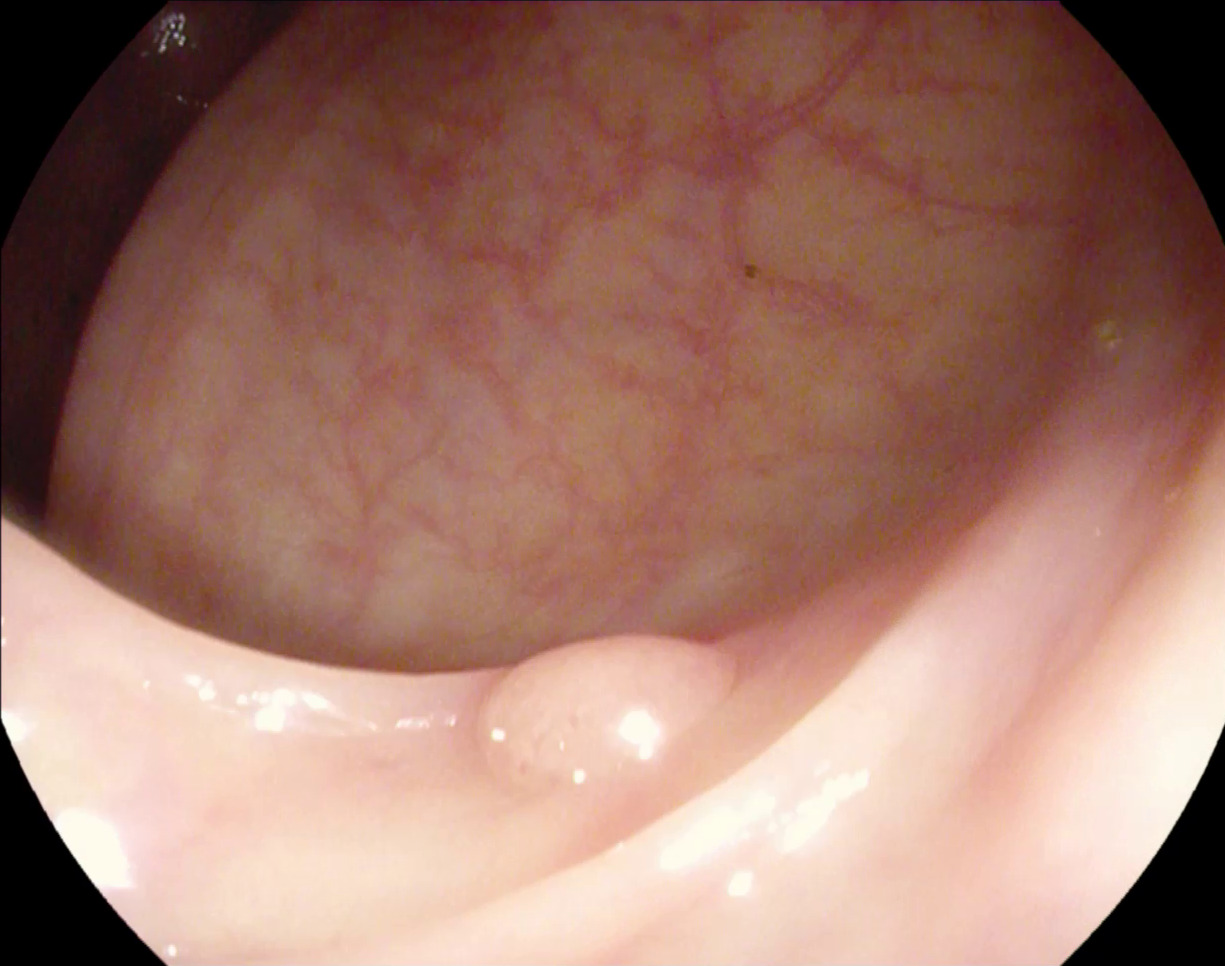} & \includegraphics[width = 1in, height = 1in]{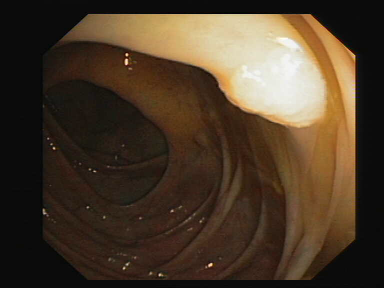} & \includegraphics[width = 1in, height = 1in]{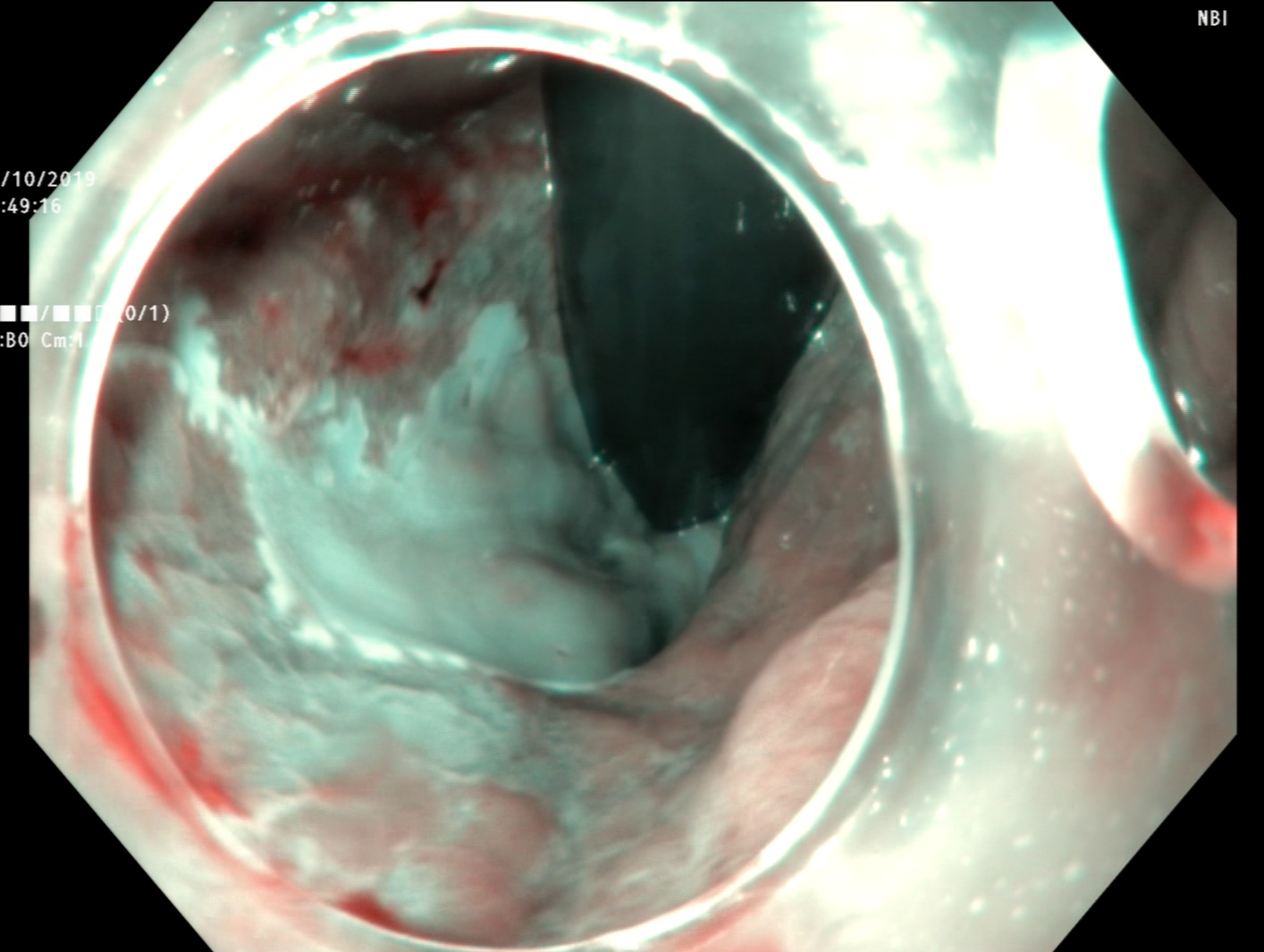} \\
               \includegraphics[width = 1in, height = 1in]{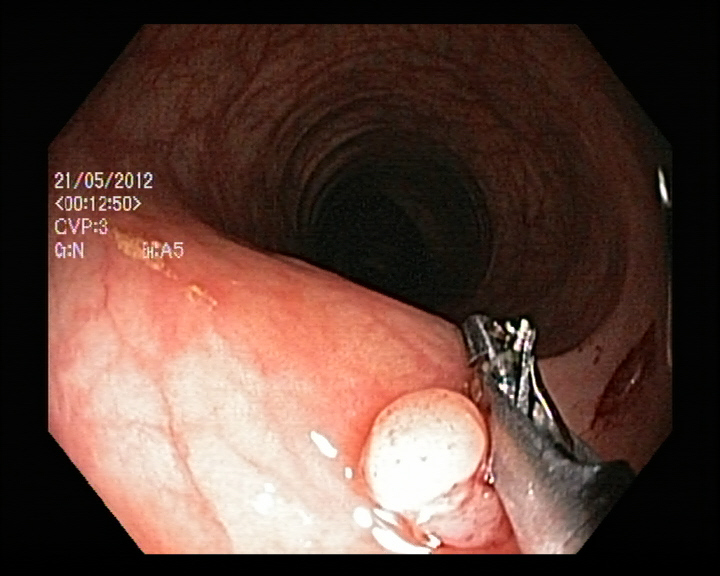} &   \includegraphics[width = 1in, height = 1in]{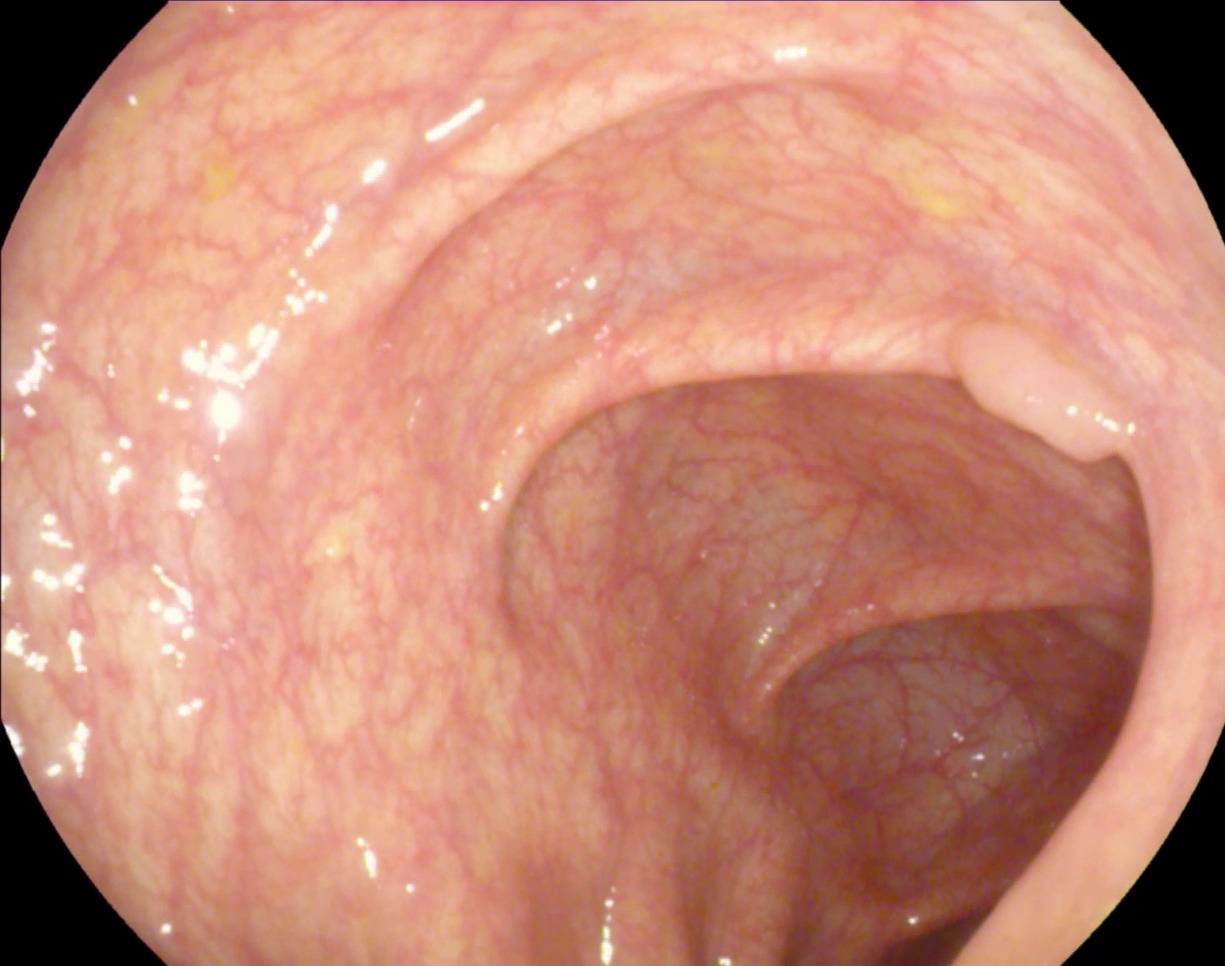} & \includegraphics[width = 1in, height = 1in]{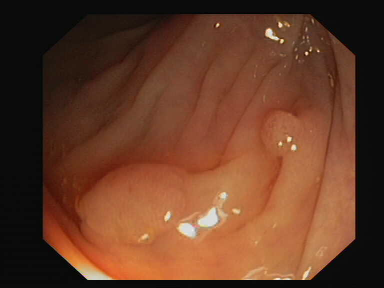} & \includegraphics[width = 1in, height = 1in]{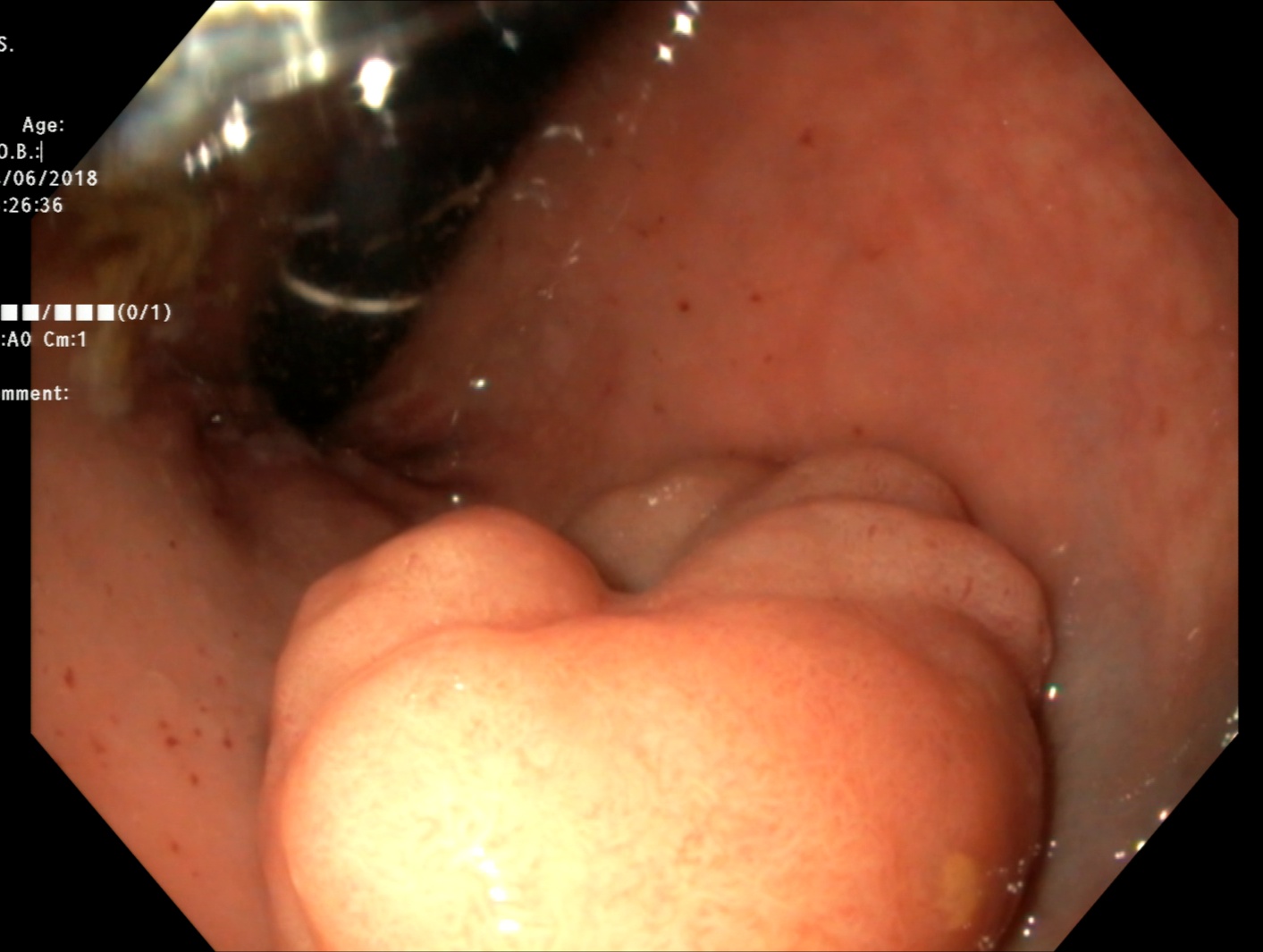} \\
                \includegraphics[width = 1in, height = 1in]{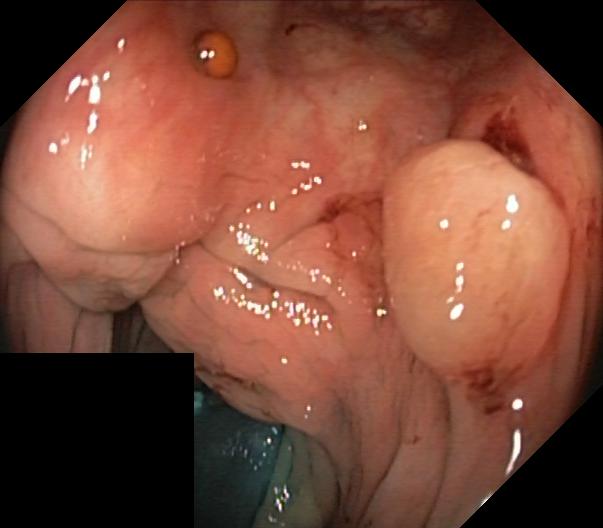} &   \includegraphics[width = 1in, height = 1in]{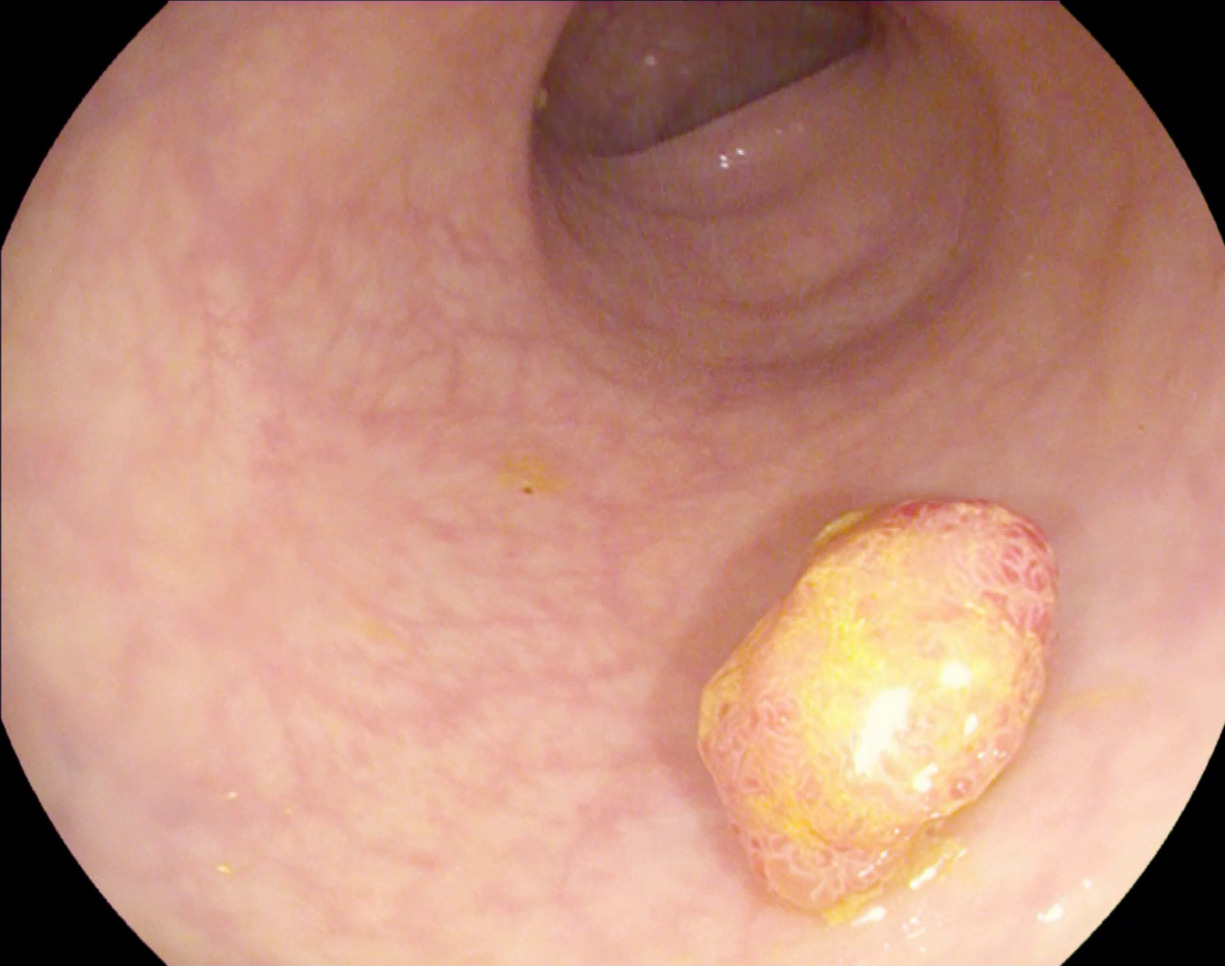} & \includegraphics[width = 1in, height = 1in]{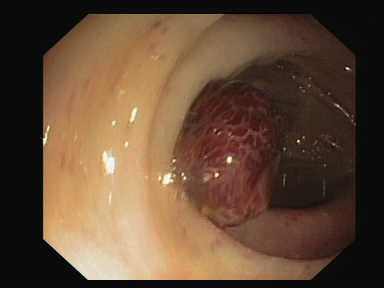} & \includegraphics[width = 1in, height = 1in]{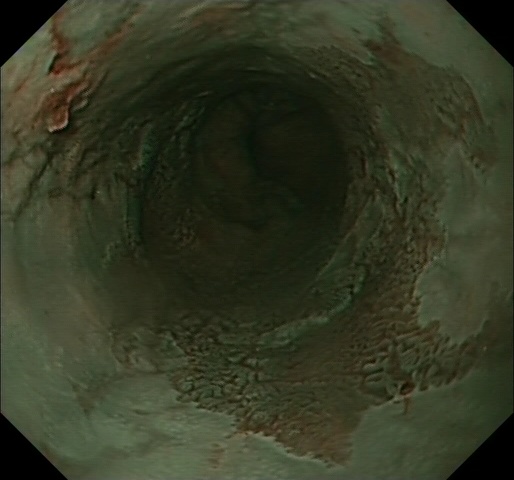} \\
                          
        \end{tabular}
        \caption{Sample images from the datasets. Kvasir-SEG (left column) was used as the training data, the remaining datasets are test-sets.}
        \label{fig:dataset_examples}
    \end{figure}

\textbf{Datasets. }
 Evaluating the the generalizability of a given predictor requires testing it on \gls{ood} data. Though this can to some extent be achieved by carefully designing stress-tests~\cite{damour2020underspecification}, a more straight-forward approach is to simply leverage existing \gls{ood} datasets. To this end, a number of polyp-segmentation datasets were selected. The names, sizes, resolutions and availabilities of these datasets is shown in~\Cref{tab:datasets}. Sample images and masks from the datasets can be seen in~\Cref{fig:dataset_examples}. Kvasir-SEG was selected as the training dataset, and partitioned into a 80/10/10 split as training/validation/test data. 
 \begin{table}[htb]
        \centering
                \caption{Dataset Overview. The training dataset is marked using "*".}

       \begin{tabularx}{\linewidth}{lXXX}
        \toprule
        Dataset & Resolution & Size & Availability \\
        \midrule
        Kvasir-SEG*~\cite{kvasir} & Variable & 1000 & Public \\
        Etis-LaribDB~\cite{etis-larib} & 1255x966 & 196  & Public \\
        CVC-ClinicDB~\cite{cvc-clinic} & 388x288 & 612  & Public \\
        EndoCV2020~\cite{endocv2020} & Variable & 127  & On Request \\
        \bottomrule
    \end{tabularx}
        \label{tab:datasets}
    \end{table}
    
\textbf{Metrics}
We used two metrics to evaluate generalizability. To evaluate raw performance, we used \gls{iou}, which is defined as follows:

\begin{equation*}
        IoU(y, \hat{y}) = \frac{\sum \{y=\hat{y}\} }{\sum \{y=1\} \cup \{\hat{y}=1\}\}}
    \end{equation*}
    
Measuring the \gls{iou} scores across all the aforementioned datasets, naturally, provides an indication of the generalizability of the given predictor. Though it is of course impossible to account for all distributional shifts that may occur in deployment, high degrees of generalization across multiple datasets should nevertheless indicate a sufficient level of generalization.

\textbf{Statistical tests}
Two different tests were used to ascertain statistical significance. In cases where the \glspl{iou} distribution was approximately normally distributed, for instance when comparing \gls{iou} samples for a given pair of models on one dataset, an independent-sample t-test was used. When comparing across models, the Mann-Whitney U-test was chosen due to the multi-modality to the resulting distributions. All p-values can be found in the supplementary material. 

\textbf{Implementation details. }
 All experiments where conduced using Nvidia Tesla-V100 GPUs. The experiments were implemented in Python 3.7.9 using PyTorch 1.8.0 and segmentation-models-pytorch~\cite{smp}. The source code as well as all of the raw data is available at \url{https://anonymous.4open.science/r/SegmentationConsistencyTraining-84EB}
 
 
 The augmentation method used both for the baseline and as part of Consistency Training was implemented using the albumentations library~\cite{albumentations}, and consisted of the following transformations: RandomRotate90, GaussNoise, ImageCompression, OpticalDistortion and ColorJitter. For the regular augmentation baseline, the augmentation probability was set to $0.5$, in which case all of the aforementioned transformations were applied. All hyperparameters can be found in the supplementary material.
 
\subsection{Out of Distribution Generalization}
\begin{figure}[htb]
    \centering
    \includegraphics[width=0.75\linewidth]{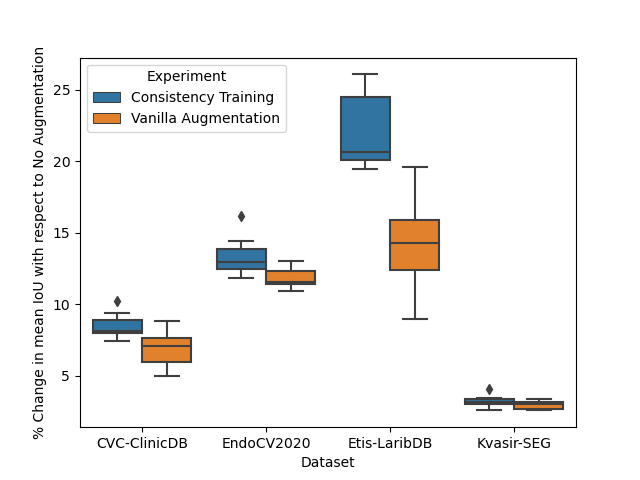}
    \caption[Consistency Training improvements]{Improvements due Consistency Training and Data Augmentation as a percentage the mean \gls{iou} without augmentation across datasets.}
    \label{fig:consistency_training_improvement}
\end{figure}

\Cref{tab:aug_ious} shows the mean \glspl{iou} for models trained with and without data augmentation, and models trained with Consistency Training. Comparing Consistency Training and conventional data augmentation for each model, statistical significance was achieved for all models except the TriUnet on the Etis-LaribDB dataset, for the FPN and Unet on the CVC-ClinicDB dataset, and for the Unet on the EndoCV2020 dataset after an independent-sample t-test. When comparing across all tested models, Consistency Training improves generalization by a statistically significant margin (p<0.01) on all \gls{ood} datasets over conventional augmentation after a Mann-Whitney U-test. A bar-plot comparing the improvement due to Consistency Training and conventional data augmentation is shown in~\Cref{fig:consistency_training_improvement}. This shows that Consistency Training can be considered a more generalizable alternative to data augmentation. 

\begin{table}[htb]
    \centering
 \caption{Mean IoUs for training methods, precision truncated to 99\% confidence. Consistency training entries with greater performance than conventional augmentation are highlighted in bold. If they are better by a statistically significant margin (p>0.99) after an independent sample two-sided t-test, they are also marked with a "*".}
    \label{tab:aug_ious}
\begin{tabularx}{\linewidth}{llXX}
\toprule
\textbf{Model} & \textbf{No Augmentation} & \textbf{Vanilla Augmentation} & \textbf{Consistency Training}\\
\toprule
\multicolumn{4}{c}{\textbf{Kvasir-SEG (In-Distribution)}}\\
\midrule
        DeepLabV3+& 0.822 & 0.850 & \textbf{0.852 }\\
        FPN& 0.822 & \textbf{0.853} & 0.852 \\
        TriUnet& 0.817 & 0.841 & \textbf{0.845} \\
        Unet& 0.828 & \textbf{0.851} & \textbf{0.851} \\
\midrule
\multicolumn{4}{c}{\textbf{Etis-LaribDB (Out of Distribution)}}\\
\midrule
        DeepLabV3+& 0.417 & 0.472 & \textbf{0.505}* \\
        FPN& 0.404 & 0.440 & \textbf{0.475}* \\
        TriUnet& 0.309 & 0.410 & \textbf{0.434} \\
        Unet& 0.403 & 0.447 & \textbf{0.481}* \\
\midrule
\multicolumn{4}{c}{\textbf{CVC-ClinicDB (Out of Distribution)}}\\
\midrule
        DeepLabV3+& 0.684 & 0.733 & \textbf{0.740} \\
        FPN& 0.675 & 0.715 & \textbf{0.727}* \\
        TriUnet& 0.623 & 0.684 & \textbf{0.696 }\\
        Unet& 0.679 & 0.717 & \textbf{0.730}* \\
\midrule
\multicolumn{4}{c}{\textbf{EndoCV2020 (Out of Distribution)}}\\
\midrule
        DeepLabV3+& 0.608 & \textbf{0.676} & \textbf{0.676} \\
        FPN& 0.600 & 0.662 & \textbf{0.673} \\
        TriUnet& 0.577 & 0.667 & \textbf{0.684} \\
        Unet& 0.598 & 0.660 & \textbf{0.676}*\\
\bottomrule
    \end{tabularx}
\end{table}

\section{Discussion and Conclusion}
In this paper, we introduced Segmentation Consistency Training, a novel training procedure for segmentation which explicitly optimizes for consistent behaviour when an input subjected to augmentation. We showed that this improves \gls{ood} generalization by a statistically significant amount across several models when compared to conventional data augmentation. Moreover, we show that Consistency Training mitigates underspecification to a greater extent than data augmentation by analyzing performance variability. 

\subsection{Limitations}\label{limitations}
The batch size was kept constant across all experiments performed in this paper. However, as it can be argued that since Consistency Training implicitly increases the batch size, the experiments should ideally be repeated across a range of batch sizes. 

Moreover, the experiments were only performed with one specific augmentation strategy. As it may be the case that the differences would be less significant if a more sophisticated strategy was used, repeating the experiment with a range of different augmentation functions and hyperparameter values is warranted.

As the experiments were only performed on polyp datasets, it can also be argued that it is uncertain whether Consistency Training has similar impacts on other segmentation tasks. 

Finally, a larger number of samples should ideally have been collected across a wider diversity of model architectures. Increasing the granularity of the findings by other means, for instance by using a greater number of \gls{ood} datasets or designing parameterized stress-tests may also be warranted in order to develop a more thorough understanding of the impact of our methods. 

\subsection{Future Work}
We plan to investigate a number of potential improvements of this framework. Consistency was for instance in this paper quantified as the symmetric difference between the expected change in the output due to augmentation and the actual change due to augmentation. This is largely agnostic to the augmentation being performed. However, it may be beneficial to take the nature of these augmentations into account. If the image is subjected to a 90 degree rotation, for instance, the prediction would following the notion of consistency as used in this work be considered perfectly consistent so long as the pixels corresponding to the polyps are rotated, and the incorrectly classified pixels remain unchanged. However, if the model instead learns to rotate all of the pixels - even those that are incorrectly classified - it may learn a more accurate representation of what constitutes consistent behavior under rotation. That means, instead of expressing inconsistency as in~\cref{inconsistency}, one can adjust the expected change term \(a\oplus y\) to \(\hat{y}\oplus \epsilon(\hat{y})\) such that also incorrect predictions can be considered consistent so long as they change in accordance to the nature of the perturbation model \(\epsilon(\cdot)\). The resulting loss function can then be expressed as:
\begin{equation*}
    \overline{\mathcal{C}}(\hat{y}, \hat{a}) = \sum \frac{\Theta(\hat{y}, \hat{a},  \hat{y}, \epsilon( \hat{y}))}{\bigcup(\hat{y}, \hat{a}, \epsilon(\hat{y}))}
\end{equation*}
Which is equivalent to:
\begin{equation*}
    \overline{\mathcal{C}}(\hat{y}, \hat{a}) = \sum \frac{\Theta(\hat{y}, \hat{a}, \epsilon( \hat{y}))}{\bigcup(\hat{y}, \hat{a}, \epsilon( \hat{y}))}
\end{equation*}
This also has the advantage of being independent of the labels themselves. This may alleviate complications that may arise as a consequence of poor and/or incomplete labeling which would otherwise affect what the models learn to associate with consistent behaviour. 

Repeating the experiments in this paper on a multitude of other segmentation tasks, for instance scene segmentation for autonomous vehicles, is also warranted. Evaluating Consistency Training through the use of stress-tests, for instance by augmenting datasets with a disjoint set of transformations as those used for training, may also provide some insights. 

Further, one could investigate whether the consistency-training framework also can be implemented in the context of classification, object detection, or other applications of Deep Learning, and if similar improvements to generalizability can be shown in other domains. 

Finally, one may compare the learned features of models trained with Consistency Training and the learned features of models trained conventionally. This could for instance be achieved through the use of Grad-CAM~\cite{gradgam} or similar methods, and may be beneficial towards determining whether the model has learned at least partial invariance to the given augmentations. 

\section{Acknowledgment}
The research presented in this paper has benefited from the Experimental Infrastructure for Exploration of Exascale Computing (eX3), which is financially supported by the Research Council of Norway under contract 270053.
\printbibliography

\section*{Checklist}

\begin{enumerate}

\item For all authors...
\begin{enumerate}
  \item Do the main claims made in the abstract and introduction accurately reflect the paper's contributions and scope?
    \answerYes{}. See \cref{fig:consistency_training_improvement} and \cref{tab:aug_ious}. 
  \item Did you describe the limitations of your work?
    \answerYes{}. See \cref{limitations}
  \item Did you discuss any potential negative societal impacts of your work?
    \answerNA{}
  \item Have you read the ethics review guidelines and ensured that your paper conforms to them?
    \answerYes{}
\end{enumerate}

\item If you are including theoretical results...
\begin{enumerate}
  \item Did you state the full set of assumptions of all theoretical results?
    \answerNA{}
        \item Did you include complete proofs of all theoretical results?
    \answerNA{}
\end{enumerate}

\item If you ran experiments...
\begin{enumerate}
  \item Did you include the code, data, and instructions needed to reproduce the main experimental results (either in the supplemental material or as a URL)?
     \answerYes{}. Code and data was made available in the form of a Github repository. 
  \item Did you specify all the training details (e.g., data splits, hyperparameters, how they were chosen)?
        \answerYes{}. See supplementary material. 
    \item Did you report error bars (e.g., with respect to the random seed after running experiments multiple times)?
    \answerYes{}. See \cref{fig:consistency_training_improvement}. If the improvement error bars are unsatisfactory, the raw data is also available on the github repository. 
    \item Did you include the total amount of compute and the type of resources used (e.g., type of GPUs, internal cluster, or cloud provider)? 
    \answerYes{}. This was discussed in \Cref{setup} and in the acknowledgment.
\end{enumerate}

\item If you are using existing assets (e.g., code, data, models) or curating/releasing new assets...
\begin{enumerate}
  \item If your work uses existing assets, did you cite the creators?
    \answerYes{}. See References list. 
  \item Did you mention the license of the assets?
    \answerNA{}
  \item Did you include any new assets either in the supplemental material or as a URL?
    \answerYes{}. The code can be found on the github repository. 
  \item Did you discuss whether and how consent was obtained from people whose data you're using/curating?
    \answerNA{}
  \item Did you discuss whether the data you are using/curating contains personally identifiable information or offensive content?
    \answerNA{}
\end{enumerate}

\end{enumerate}

\end{document}


\maketitle

\section{ Non-weighted Consistency Training}\label{non_weighted_ctraining}

\begin{figure}[htb]
    \centering
    \includegraphics[width=\linewidth]{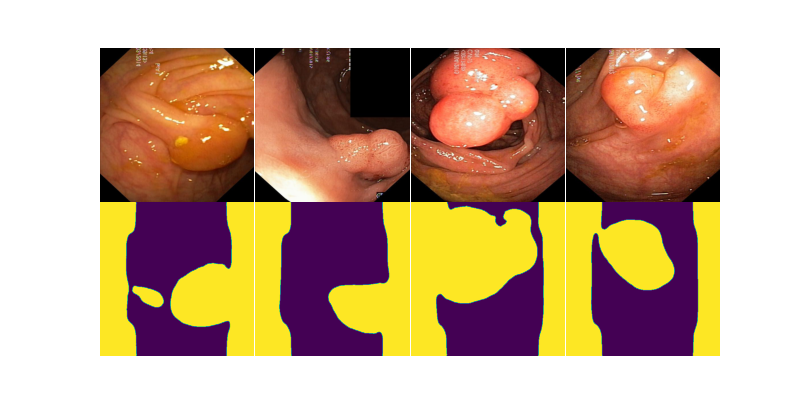}
    \caption[Unweighted Consistency example]{When the Inconsistency term is not modulated dynamically, the model can quickly learn to predict artifacts around the edges of the image. As polyps can rarely be found in these regions, the consistency term is minimized by predicting consistently wrong predictions where there typically are not polyps. To avoid instability in this regard, dynamic weighing is necessary. }
    \label{fig:non_weighted_ctraining}
\end{figure}

An effect of non-weighted consistency training is presented in Figure~\ref{fig:non_weighted_ctraining}. The lack of weighting can sometimes induce instability; the model may learn to prioritize consistency at the cost of segmentation accuracy. In \Cref{fig:non_weighted_ctraining}, the model has learned to consistently predict positively along the margins of the image. As polyps are rarely found in these regions in the training data, the search will be biased toward predicting positively in these regions, as this will minimize the overall loss due to the reduced contribution from the consistency term. It should be noted, however, that these artifacts will gradually disappear given sufficient training. Nevertheless, the weighting scheme permits more efficient and stable training.  

\section{Statistical Tests}
Two different tests were used to ascertain statistical significance. In cases where the \glspl{iou} distribution was approximately normally distributed, for instance when comparing \gls{iou} samples for a given pair of models on one dataset, an independent-sample t-test was used. When comparing across models, the Mann-Whitney U-test was chosen due to the multi-modality to the resulting distributions.
\begin{table}[H]
    \centering
    \begin{tabularx}{\linewidth}{lXr}
            \toprule
            Dataset & U-Statistic & p-Value \\
            \midrule
            Kvasir-SEG & 725.0 & 0.23673 \\
            Etis-LaribDB & 388.0 & 0.00004\\
            CVC-ClinicDB & 491.0& 0.00150 \\
            EndoCV2020 & 411.0 & 0.00009    \\
            \bottomrule
        \end{tabularx}
        \caption{Results from a Mann-Whitney U-test for each dataset when comparing the \glspl{iou} for Consistency Training vs conventional data augmentation across models}
        \label{tab:ttest_avgs_consistency}
    \end{table}
    
        \begin{table}[H]
    \centering
    \begin{tabularx}{\linewidth}{lXXXX}
    \toprule
      Model & CVC-ClinicDB & EndoCV2020 & Etis-LaribDB & Kvasir-SEG\\
      \midrule
      DeepLabV3+       & 0.029 & 0.901 & \textbf{0.003} & 0.444\\
      FPN           & \textbf{0.004} & 0.038 & \textbf{0.005} & 0.939\\
      TriUnet       & 0.211 & 0.024 & 0.141 & 0.330\\
      Unet          & \textbf{0.000} & \textbf{0.001} & \textbf{0.006} & 0.899\\
      \bottomrule
    \end{tabularx}
    \caption[T-test results consistency training]{p-values for each model and dataset between the \glspl{iou} of the given models trained with consistency training versus when trained with data augmentation}
    \label{tab:ttest_per_dataset_consistency}
\end{table}

\section{Hyperparameters}

\begin{table}[htb]
    \centering
    \caption{Overview of augmentation functions and corresponding non-default hyperparameters used in this work.}
\begin{tabularx}{\textwidth}{XX}
    \toprule
    \textbf{Invariance} & \textbf{Albumentation Function}\\
    \midrule
    Perspective &Flip() \newline RandomRotate90()\\
    \midrule
    Image quality &GaussNoise(max=0.01) \newline ImageCompression(max=100, min=10)\\
    \midrule
    Camera models&OpticalDistortion(distort\_limit=10) \\
    \midrule
    Lighting conditions & ColorJitter(brightness=0.2,\newline hue=0.2, contrast=0.2, \newline saturation=0.2) \\
    \bottomrule
\end{tabularx}
    \label{tab:vanilla_aug}
\end{table}

 \begin{table}[htb]
        \centering
        \caption{Hyperparameters and objects used to train all models.}
        \begin{tabularx}{\linewidth}{llX}
        \toprule
        Component & Type & Hyperparameters \\
        \midrule
        Dataloader & - & \(batch\_size = 8\) \\
        && \(\hbox{train/val/test split} = 80/10/10\)\\
        \midrule
        Optimizer & Adam & \(lr = 0.00001\)\\
        \midrule
        Scheduler & Cosine Annealing w/ Warm Restarts & \(T_0=50\) \\
        & & \(T_{mult}=2\) \\
        \midrule
        Evaluation & Loss-based Early Stopping & \(epochs=300\)\\
        \bottomrule
        \end{tabularx}
            \label{tab:hyperparameters}
\end{table}